\documentclass[10pt,logo,copyright]{nvidiatechreport}
\usepackage{subcaption}
\usepackage{wrapfig}
\usepackage{makecell}
\usepackage{multirow}

\usepackage[numbers, sort]{natbib}
\definecolor{citecolor}{HTML}{0071BC}
\definecolor{linkcolor}{HTML}{ED1C24}

\usepackage{nicefrac}
\usepackage{float}
\usepackage{array}

\usepackage[capitalize]{cleveref}
\crefname{section}{\S}{\S\S}
\crefname{subsection}{\S}{\S\S}
\crefname{table}{\text{Tab.}}{\text{Tab.}}
\Crefname{table}{Table}{Tables}
\crefname{figure}{\text{Fig.}}{\text{Fig.}}
\Crefname{figure}{Figure}{Figures}
\crefname{equation}{\text{Eq}}{\text{Eq}}

\newcommand{\ours}{$\gamma$-World\xspace}
\newcommand{\ourslong}{Gamma-World\xspace}

\title{\ourslong: Generative Multi-Agent World Modeling Beyond Two Players}

\author{
  \textbf{Fangfu Liu}$^{1,2*}$ \quad
  \textbf{Kai He}$^{1,3,4*}$ \quad
  \textbf{Tianchang Shen}$^{1}$ \quad
  \textbf{Tianshi Cao}$^{1}$ \quad
  \textbf{Sanja Fidler}$^{1,3,4}$ \quad
  \textbf{Yueqi Duan}$^{2}$ \quad
  \textbf{Jun~Gao}$^{1}$  \newline
  \textbf{Igor Gilitschenski}$^{3,4\dagger}$ \quad
  \textbf{Zian Wang}$^{1\dagger}$ \quad
  \textbf{Xuanchi Ren}$^{1\dagger}$ \\
  \small $^{1}$NVIDIA \quad $^{2}$Tsinghua University \quad $^{3}$University of Toronto \quad $^{4}$Vector Institute
}

\fancypagestyle{firststyle}{%
  \fancyhead[L]{\includegraphics[width=90pt]{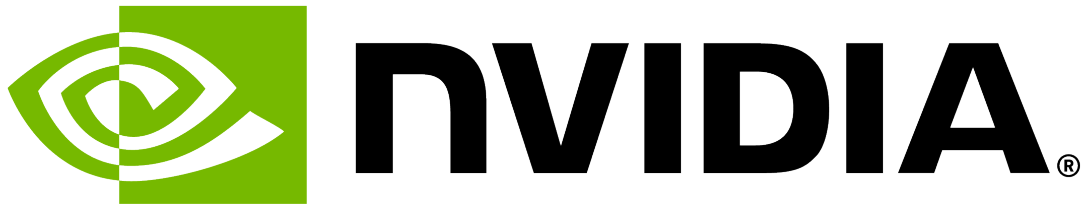}}%
  \fancyhead[R]{{\footerfont\itshape\monthyeardate\today}}%
  \fancyhead[C]{}%
  \fancyfoot[L]{%
    \footerfont$^{*}$Equal contribution; order interchangeable.\\
    \footerfont$^{\dagger}$Joint advising.
  }%
  \fancyfoot[R]{\copyrightext}%
  \fancyfoot[C]{\footerfont\bfseries\relax}%
}

\begin{document}
\maketitle
\begin{center}
\vspace{-1em}
\textbf{Project page:} 
\href{https://research.nvidia.com/labs/sil/projects/gamma-world}{research.nvidia.com/gamma-world}
\end{center}
\begin{figure}[ht!]
  \centering
  \includegraphics[width=\linewidth]{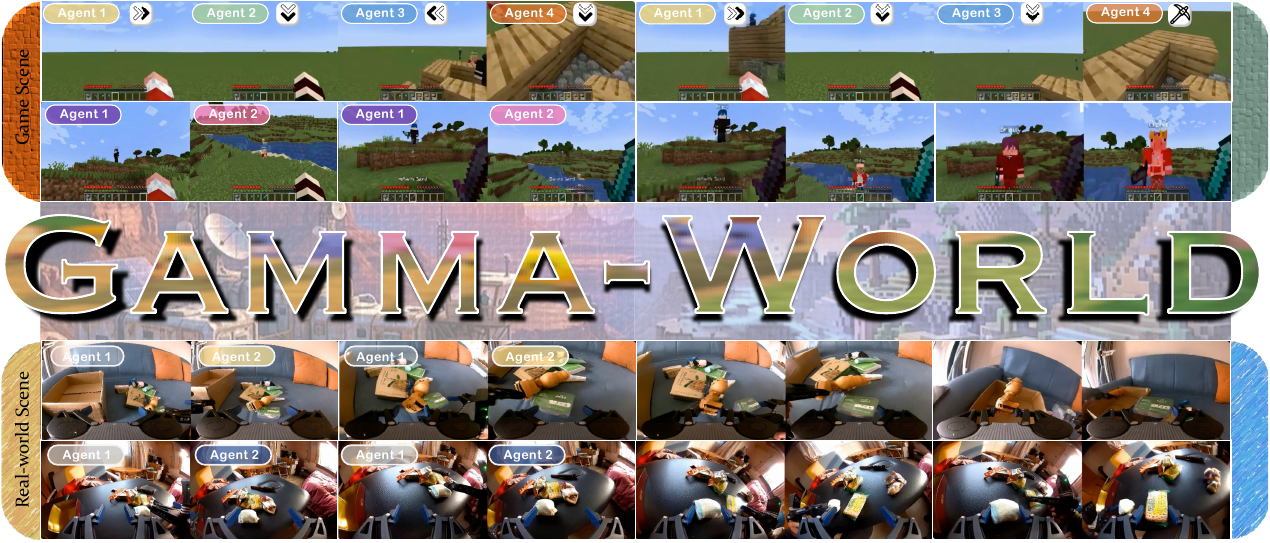}
  \caption{We propose \ours, a novel generative multi-agent world model from virtual games to real-world environments. More results and video demos are available on our \href{https://research.nvidia.com/labs/sil/projects/gamma-world}{project page}.}
  \label{fig:teaser}
\end{figure}

\begin{abstract}
World models for interactive video generation have largely focused on single-agent settings, where future observations are generated from a single control signal. 
However, many generated environments require multi-agent interaction: multiple players, robots, or embodied agents act simultaneously within a shared space. 
Scaling world models to such settings requires a principled multi-agent design: agents should remain independently controllable, permutation-symmetric, and support efficient inference while maintaining consistency across time and perspectives.
In this paper, we present \ours, a generative multi-agent world model for interactive simulation. 
\ours introduces \emph{Simplex Rotary Agent Encoding}, a parameter-free extension of 3D RoPE that represents agents as vertices of a regular simplex in rotary angle space. This gives each agent a distinct phase while making all agents permutation-equivalent, enabling scalable agent identity without learned per-slot identities or a fixed agent ordering. 
To avoid dense all-to-all attention across agents, we further propose \emph{Sparse Hub Attention}, where learnable hub tokens mediate token-interaction across agents, reducing cross-agent attention cost from quadratic to linear in the number of agents. 
For real-time rollout, we distill a full-context diffusion teacher into a causal student that generates temporal blocks sequentially with KV caching, enabling action-responsive generation at 24 FPS.
Experiments in multiplayer virtual environments show that \ours improves video fidelity, action controllability, and inter-agent consistency over slot-based and dense-attention baselines, while generalizing from two to four players without additional training. 

\end{abstract}

\abscontent

\section{Introduction}
\label{sec:introduction}

The worlds we wish to simulate are populated, not solitary.
Players cooperate and compete in the same game, robot arms coordinate around shared objects, and embodied agents act under mutual physical and visual constraints.
Controllable multi-agent world modeling is therefore a necessary step toward multiplayer game generation~\cite{bruce2024genie,savva2026solaris}, interactive simulation~\cite{ren2025cosmos-drive-dreams,ali2025world-simulation}, and embodied AI~\cite{fung2025embodied-ai-agents,feng2026multi-agent-embodied-ai}.

Despite this need, most video world models~\cite{alonso2024diffusion,valevski2024diffusion,he2025matrix,xiao2025worldmem,hyworld2025,worldplay2025,yu2025cam} remain single-agent simulators: they roll out future observations conditioned on one action stream, one user input, or one controllable viewpoint.
Moving from single-agent to multi-agent simulation raises a new consistency requirement on top of standard video generation: generated observations must be consistent not only across time, but also across agent perspectives, since all agents share and act upon the same evolving world.

A concurrent effort in this direction is Solaris~\cite{savva2026solaris}, which builds a multiplayer Minecraft world model by combining a dense joint attention block over all agent tokens with a learned per-player ID embedding.
While effective in the two-player setting, this design has two structural limitations.
First, dense joint attention couples every agent token to every other agent token, so its cost grows quadratically with the number of agents, and becomes increasingly restrictive for real-time rollouts beyond two players.
Second, agents in a shared world are intrinsically exchangeable: two agents with identical capabilities should not be treated differently simply because they occupy different slots.
A learned per-slot ID embedding violates this symmetry and ties the model to a fixed player roster that cannot be extended without retraining.
This leaves open the question of whether a video world model can represent multiple agents in a way that is individually controllable, permutation-symmetric, and scalable beyond two players.

We address this question with \ours, a generative multi-agent world model for interactive simulation that scales beyond two players.
\ours extends standard 3D RoPE with an explicit agent axis through \emph{Simplex Rotary Agent Encoding}.
Rather than assigning agents scalar indices or learned identity vectors, we place them at the vertices of a regular simplex in rotary angle space.
Geometrically, this puts all agents at equal pairwise distances, so every pair is permutation-equivalent while each agent retains a distinct rotary phase.
The encoding is parameter-free, does not rely on a fixed learned slot identity, and can be instantiated for different numbers of agents without changing the video transformer architecture.

To complement this symmetric agent representation with efficient cross-agent interaction, we introduce \emph{Sparse Hub Attention}.
Within each causal block, agent tokens attend to their own stream and to a small set of learnable hub tokens.
The hub tokens aggregate information across agents and broadcast it back, providing a shared communication pathway without dense pairwise attention.
This hub-mediated topology preserves a shared communication pathway among agents without dense pairwise interaction, reducing the dominant cross-agent cost from quadratic to linear in the number of agents and substantially relieving the scaling pressure of going beyond two players.
To make the model deployable as a real-time interactive simulator, we further distill a bidirectional multi-agent teacher into a block-causal student with KV caching, enabling 24-FPS streaming autoregressive rollouts that respond to newly issued actions.

We evaluate \ours on multiplayer virtual environments with two and four players spanning movement, mining, combat, and building scenarios.
Across these settings, \ours improves video fidelity, action controllability, and inter-agent consistency over slot-based and dense-attention baselines. 
Ablations validate the contributions of simplex rotary encoding and sparse hub communication. Scaling studies further show that, benefiting from the permutation-symmetric simplex agent encoding, the model generalizes from two to four players without additional training. 

In summary, our main contributions are:
\begin{itemize}
    \setlength{\itemsep}{0pt}
    \setlength{\parsep}{0pt}
    \setlength{\topsep}{0pt}
    \item We identify permutation symmetry as a fundamental property of multi-agent world models and propose \emph{Simplex Rotary Agent Encoding}, a parameter-free rotary identity encoding that preserves permutation symmetry while maintaining distinct agent identities.
    \item We propose \emph{Sparse Hub Attention}, a cross-agent communication mechanism that reduces cross-agent attention cost from quadratic to linear in the number of agents.
    \item We demonstrate effective multi-agent simulation in multiplayer environments with two and four players, with ablations validating the proposed agent encoding and efficiency design.
\end{itemize}

\section{Related Work}

\noindent\textbf{Video generation.}
Diffusion-based generative models~\cite{ho2020denoising, lipman2022flow, song2020score} have become a leading approach for video generation.
Built upon latent diffusion models~\cite{rombach2022high}, recent methods~\cite{chen2024videocrafter2, guo2023animatediff, yang2024cogvideox} perform generative modeling in the latent space, enabling more efficient synthesis of high-quality videos.
In parallel, autoregressive video generation approaches~\cite{chen2024diffusion, henschel2025streamingt2v, kim2024fifo} have attracted increasing attention for their potential to extend generation to arbitrary-length sequences, thereby offering a promising basis for building world models.
Driven by scalable architectures~\cite{peebles2023scalable} and increasingly sophisticated data curation pipelines, large-scale video generation systems~\cite{veo, sora, wan, kling, gao2025seedance, kong2024hunyuanvideo} trained on large-scale datasets have further exhibited emergent zero-shot abilities to understand, model, and manipulate the visual world~\cite{wiedemer2025video}, bringing physical world simulation closer to reality.

\noindent\textbf{Video world models.}
With the development of video diffusion transformers and the rapid advances in text-to-video and image-to-video generation, a growing body of world-modeling research has begun to repurpose video diffusion models as visual simulators~\cite{yang2023unisim,bar2025navigation,liang2024dreamitate}.
Rather than learning compact abstract states for downstream planning and decision making, these methods directly model future visual observations through generative video prediction.
We therefore refer to this family of approaches as video world models, highlighting their central reliance on video generation for simulating the evolution of the visual world.
Recent video world models have been increasingly studied in embodied scenarios, where agents must anticipate the visual consequences of actions and interactions.
Representative applications span robotics~\cite{mereu2025generative,yang2023learning,li2025unified,li2026causal,ye2026world,gao2026dreamdojo}, video games~\cite{alonso2024diffusion,he2025matrix,valevski2024diffusion,xiao2025worldmem, fan2022minedojo}, autonomous driving~\cite{agarwal2025cosmos,hu2023gaia}, and physical simulation~\cite{li2025pisa,yuan2026inference}, suggesting their potential as general-purpose simulators for dynamic visual environments.
Beyond open-loop video rollout, recent efforts~\cite{hyworld2025,worldplay2025,yu2025cam} further move toward interactive and temporally consistent world modeling, where the simulated environment can react to user or agent actions while preserving coherent scene structure, object dynamics, and long-range temporal consistency.

\noindent\textbf{Video diffusion model distillation.}
Real-time generation is a crucial requirement for building world models.
A common strategy for video diffusion models is to accelerate sampling through distillation~\cite{salimans2022progressive, geng2025mean, frans2024one}, which enables few-step inference and thus supports real-time synthesis.
Several works~\cite{sauer2024fast, sauer2024adversarial, kang2024distilling, lin2025diffusion, lin2024sdxl, lin2025autoregressive} introduce adversarial training objectives to reduce the number of denoising steps, but such methods are often prone to optimization instability and model collapse.
Another line of approaches~\cite{yin2024one, yin2024improved, lu2025adversarial} leverages variational score distillation~\cite{wang2023prolificdreamer} to obtain strong few-step generation performance.
Beyond these general acceleration techniques, CausVid~\cite{yin2025slow} tackles real-time autoregressive generation by transferring knowledge from a bidirectional diffusion teacher to a causal student model.
Building on this causal generation setup, Self-Forcing~\cite{huang2025self} further improves rollout training to reduce the exposure bias accumulated during long-horizon generation.

\section{Method}

We consider the problem of synchronized action-conditioned multi-agent video generation. Formally, the goal is to learn a function $\gamma\text{-World}(\{o_{1:t}^p\}_{p=1}^P,\, \{a_{1:t}^p\}_{p=1}^P)$, that takes a sequence of past observations and actions for $P$ agents as input and produces the next observation for each agent $\{o_{t+1}^p\}_{p=1}^P$. Because we focus on the multi-agent setting, it is important to note that $\{o_t^p\}_{p=1}^P$ correspond to different perspectives of the same underlying world state at time $t$.

\begin{figure*}[t]
  \centering
  \includegraphics[width=\linewidth]{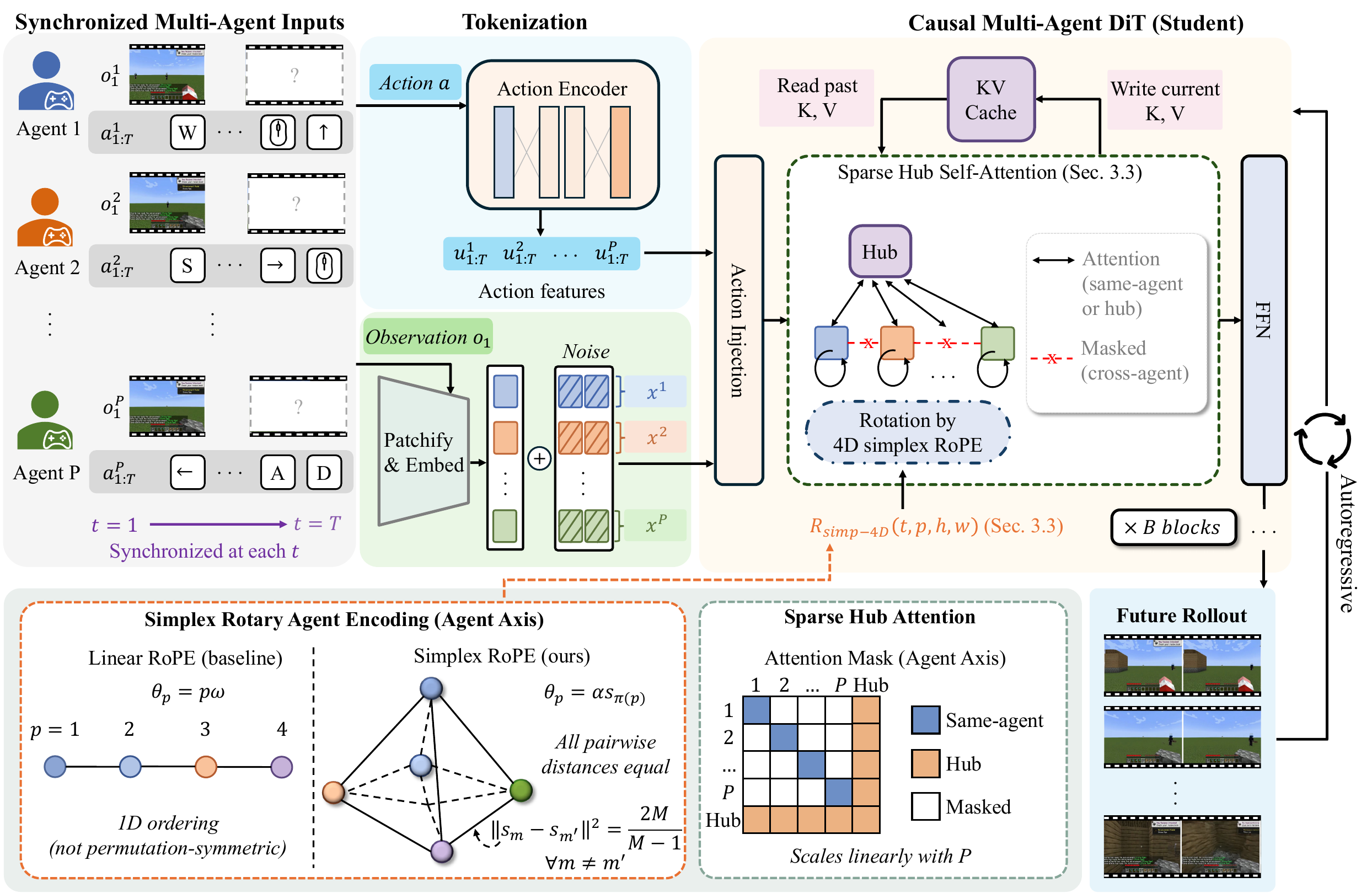}
  \caption{\textbf{Method overview.}
  \ours takes synchronized observations and actions from multiple agents as input, tokenizes each agent stream with shared visual and action encoders, and generates future multi-agent rollouts with a causal multi-agent DiT.
  We formulate the input with an explicit synchronized agent axis, encode exchangeable agent identity using Simplex Rotary Agent Encoding (\S\ref{sec:simplex-rotary-agent-encoding}), and route cross-agent information through Sparse Hub Attention (\S\ref{sec:sparse-hub-attention}).
  During streaming inference, the causal student uses KV caches for past visual tokens and hub states to preserve block-causal generation while scaling efficiently with the number of agents (\S\ref{sec:causal-distillation}).}
  \label{fig:method}
\end{figure*}

Our proposed model builds on transformer-based latent video diffusion models adapted for autoregressive generation, in which rotary position embeddings encode spatial and temporal locations (\S\ref{sec:preliminaries}). We modify this position embedding to account for agent identities and propose a multi-agent aware attention masking mechanism to reduce computational cost (\S\ref{sec:simplex-rotary-agent-encoding}). Finally, the full model is trained in two steps: first, a bidirectional teacher model is trained, and then it is distilled into a causal student model that supports the streaming setting (\S\ref{sec:causal-distillation}). An overview of the method is provided in figure~\ref{fig:method}.

\subsection{Preliminaries}
\label{sec:preliminaries}

\noindent\textbf{Latent video diffusion.}
We build on DiT-based latent video diffusion transformers~\cite{peebles2023scalable,wan}.
Let $\mathbf{z}_0\in\mathbb{R}^{T\times H\times W\times C_z}$ denote a clean video latent with $T$ temporal positions, $H\times W$ spatial positions, and $C_z$ channels per token.
We adopt the flow-matching objective~\cite{lipman2022flow}: given $\mathbf{z}_0$, a noise sample $\boldsymbol{\epsilon}\sim\mathcal{N}(\mathbf{0},\mathbf{I})$, and a noise level $\sigma\in[0,1]$, we form the linear interpolant
\begin{equation}
\mathbf{z}_{\sigma}=(1-\sigma)\,\mathbf{z}_0+\sigma\,\boldsymbol{\epsilon},
\end{equation}
and train a velocity field $v_\theta$ to regress its time derivative $\boldsymbol{\epsilon}-\mathbf{z}_0$:
\begin{equation}
\mathcal{L}_{\mathrm{FM}}
=
\mathbb{E}_{\mathbf{z}_0,\,\boldsymbol{\epsilon},\,\sigma}\!\left[
\left\|v_{\theta}(\mathbf{z}_{\sigma},\sigma,\mathcal{C})-(\boldsymbol{\epsilon}-\mathbf{z}_0)\right\|_2^2
\right],
\end{equation}
where $\mathcal{C}$ denotes conditioning signals such as initial observations and agent actions.

\noindent\textbf{Causal autoregressive video generation.}
Standard video diffusion transformers are typically bidirectional: during denoising, tokens can attend to the full latent video, which improves global consistency but prevents streaming rollout.
To enable autoregressive generation, Self-Forcing~\cite{huang2025self} combines Diffusion Forcing~\cite{chen2024diffusion} with block-causal attention.
The latent sequence is partitioned into temporal blocks of $n$ frames, an independent noise level $\sigma_b\sim\mathcal{U}(0,1)$ is sampled for each block, and each query attends only to keys from the same or earlier blocks.
Attention remains bidirectional within each block, matching streaming inference where the model repeatedly denoises the current block conditioned on previously generated blocks.

\noindent\textbf{Rotary Position Embedding (RoPE).}
Video diffusion transformers commonly use 3D RoPE~\cite{rope} to inject relative position into self-attention by rotating query and key features along factorized temporal and spatial axes.
The rotary head dimensions are split into temporal, height, and width bands of sizes $d_t,d_h,d_w$, with $d_t+d_h+d_w=d_{\mathrm{rope}}$.
For a token at coordinate $(t,h,w)$, the 3D RoPE operator is
\begin{equation}
\mathbf{R}_{\mathrm{3D}}(t,h,w)
=
\mathrm{diag}\!\left(\mathbf{R}_{t}(t),\,\mathbf{R}_{h}(h),\,\mathbf{R}_{w}(w)\right),
\end{equation}
where each $\mathbf{R}_x(x)$ is a block-diagonal of 2D rotations whose angles follow the standard RoPE frequency schedule~\cite{rope}.



\subsection{Model Design}
\label{sec:model-design}

Our model starts by performing visual tokenization, after which a clean multi-agent latent is represented as
$\mathbf{Z}_0\in\mathbb{R}^{P\times T\times H\times W\times C_z}$,
which extends the single-agent latent $\mathbf{z}_0$ from \S\ref{sec:preliminaries} with an explicit agent axis indexed by $p\in\{1,\ldots,P\}$.
The model is conditioned on initial observations and per-agent action sequences, and predicts future latent observations for all agents jointly.
This formulation encourages generated rollouts to remain consistent across both time and agent perspectives.
At inference, the first observations $\{o_1^p\}_{p=1}^P$ are encoded as context, while future latent tokens are initialized from noise and denoised block by block under the per-agent action sequences.

\noindent\textbf{Action conditioning.}
\label{sec:action-conditioning}
Each agent is controlled by its own action sequence $\mathbf{a}^p_{1:T}$.
We use a shared action encoder $f_a$ across all agents to map each action to a hidden action feature
$\mathbf{u}_t^p = f_a(a_t^p) \in \mathbb{R}^{D}$,
where $D$ is the DiT hidden dimension.
For discrete or multi-component controls, we embed each action component and combine the embeddings with a lightweight MLP.

Let $\mathbf{x}_{\ell,p,t,h,w}\in\mathbb{R}^{D}$ denote the transformer state at layer $\ell$ for agent $p$, frame $t$, and spatial position $(h,w)$.
At transformer block $\ell$, the action feature is projected to a layer-specific action bias
\begin{equation}
\boldsymbol{\beta}_{\ell,t}^p = g_\ell(\mathbf{u}_t^p) \in \mathbb{R}^{D},
\end{equation}
and broadcast to all spatial tokens of the corresponding agent and frame:
\begin{equation}
\mathbf{x}_{\ell,p,t,h,w}
\leftarrow
\mathbf{x}_{\ell,p,t,h,w}
+
\boldsymbol{\beta}_{\ell,t}^p .
\end{equation}
The biased tokens are then passed to self-attention.
The action encoder and projection layers are shared across agents, so the same action has the same representation regardless of agent identity.

\noindent\textbf{Simplex Rotary Agent Encoding.}
\label{sec:simplex-rotary-agent-encoding}
To distinguish agents without imposing an arbitrary slot ordering, we extend the 3D RoPE coordinate from \S\ref{sec:preliminaries} with an explicit agent axis $p$.
We partition the rotary head dimension as
$d_{\mathrm{rope}}=d_t+d_p+d_h+d_w$,
which yields the 4D rotary operator
\begin{equation}
\mathbf{R}_{\mathrm{4D}}(t,p,h,w) = \mathrm{diag}\!\left(\mathbf{R}_{t}(t),\,\mathbf{R}_{p}(p),\,\mathbf{R}_{h}(h),\,\mathbf{R}_{w}(w)\right).
\end{equation}
A natural instantiation of $\mathbf{R}_{p}$ is to assign each agent a scalar phase $\boldsymbol{\theta}_p=p\,\boldsymbol{\omega}$ for some frequency vector $\boldsymbol{\omega}$.
However, this places exchangeable agents on a one-dimensional line: different pairs receive different rotary distances depending on $|p-q|$, and the indexing convention can make certain slots structurally special.
Learned per-slot embeddings have a different failure mode: they tie agent identity to a fixed roster and break permutation symmetry.
We instead seek an encoding that distinguishes agents while making all agent identities \emph{equidistant and exchangeable}.

To achieve this, we represent agents as vertices of a regular simplex in rotary angle space.
Let $V$ denote a fixed simplex pool size chosen at training time, corresponding to the maximum number of supported agent identities, with $V\leq d_p/2+1$.
We construct $V$ simplex vertices in the $d_p/2$-dimensional agent-angle space:
\begin{equation}
\mathbf{s}_v
=
\sqrt{\tfrac{V}{V-1}}\;
\mathbf{Q}\!\left(\mathbf{e}_v-\tfrac{1}{V}\mathbf{1}\right)
\in\mathbb{R}^{d_p/2},
\qquad v=1,\ldots,V,
\end{equation}
where $\mathbf{e}_v\in\mathbb{R}^{V}$ is the $v$-th one-hot vector, $\mathbf{1}\in\mathbb{R}^{V}$ is the all-ones vector, and $\mathbf{Q}$ is a linear isometry from the zero-mean subspace of $\mathbb{R}^{V}$ into $\mathbb{R}^{d_p/2}$.

The resulting vertices have unit norm and equal pairwise distance:
\begin{equation}
\|\mathbf{s}_v\|_2=1,\qquad
\|\mathbf{s}_v-\mathbf{s}_{v'}\|_2^2=\tfrac{2V}{V-1}
\quad\text{for all }v\neq v',
\end{equation}
with the proof given in Appendix~\ref{app:simplex-equidistance}.

For a batch with $P\leq V$ active agents, we sample an injective assignment
$\pi:\{1,\ldots,P\}\rightarrow\{1,\ldots,V\}$
and assign agent $p$ to simplex vertex $\mathbf{s}_{\pi(p)}$.
The agent-band rotation angles are
\begin{equation}
\boldsymbol{\theta}_p=\alpha\,\mathbf{s}_{\pi(p)},
\end{equation}
where $\alpha>0$ controls the strength of agent separation.
Replacing $\mathbf{R}_{p}$ with the corresponding simplex rotation $\mathbf{R}_{\mathrm{simp}}(\pi(p))$ gives
\begin{equation}
\mathbf{R}_{\mathrm{simp\text{-}4D}}(t,p,h,w)
=
\mathrm{diag}\!\left(
\mathbf{R}_{t}(t),
\mathbf{R}_{\mathrm{simp}}(\pi(p)),
\mathbf{R}_{h}(h),
\mathbf{R}_{w}(w)
\right).
\end{equation}

This encoding is parameter-free, permutation-symmetric, and does not introduce learned per-slot identities.
The simplex pool provides a direct mechanism for agent-count scaling.
During training, active agents are randomly assigned to distinct vertices from the fixed pool, discouraging slot-specific overfitting.
At inference, additional agents can be activated by selecting additional unused vertices from the same pool, without changing the transformer architecture or introducing new learned identities.
For pretrained video DiTs that lack an explicit agent band, we follow ReRoPE~\cite{li2026rerope} and allocate $d_p$ dimensions from the low-frequency end of the temporal band, leaving the high-frequency temporal and spatial bands intact.

\noindent\textbf{Sparse Hub Attention.}
\label{sec:sparse-hub-attention}
A direct way to model cross-agent interaction is dense joint attention over all agent tokens, as in Solaris~\cite{savva2026solaris}.
With $L=HW$ tokens per frame and $P$ agents, dense cross-agent attention over a block of $n$ frames costs $\mathcal{O}(P^2n^2L^2)$.
This becomes expensive as the number of agents grows, and it is often unnecessary: in many shared-world settings, agents influence one another primarily through a compact evolving environment state rather than through dense token-level pairwise exchange at every layer.

We therefore introduce \emph{Sparse Hub Attention} (SHA), a hub-mediated attention topology that adds a small set of learnable hub tokens as a compact shared communication state.
Agent tokens attend only to tokens from the same agent stream and to the hub tokens.
Hub tokens attend to all agents and to other hub tokens.
Direct attention between distinct agent streams is masked, so cross-agent information flows through a two-hop path: agent~$\rightarrow$~hub~$\rightarrow$~agent.

We organize the sequence as $PTL$ agent tokens followed by $TK$ hub tokens, with $K$ hub tokens per latent frame.
The hub tokens are drawn from a learnable matrix $\mathbf{H}\in\mathbb{R}^{K\times D}$, broadcast across frames, and removed from the output, so they act purely as internal communication states.
Let $\rho(i)\in\{1,\ldots,P,\mathrm{hub}\}$ denote the identity of token $i$.
The hub-and-spoke topology is defined by
\begin{equation}
\mathcal{M}_{\mathrm{hub}}(i,j)
=
\mathbf{1}\!\left[
\rho(i)=\rho(j)
\;\vee\;
\rho(i)=\mathrm{hub}
\;\vee\;
\rho(j)=\mathrm{hub}
\right].
\end{equation}
For causal autoregressive generation, we compose this topology with the block-causal mask (\S\ref{sec:preliminaries}):
\begin{equation}
\mathcal{M}(i,j)
=
\mathbf{1}\!\left[b(j)\leq b(i)\right]
\cdot
\mathcal{M}_{\mathrm{hub}}(i,j),
\end{equation}
where $b(i)$ is the temporal block index of token $i$.
The first factor enforces block-level causality, while the second enforces hub-mediated cross-agent communication.
Hub tokens reuse the temporal RoPE phase of their associated frame and use identity rotations on the agent, height, and width bands, keeping them temporally aligned while remaining neutral to agent identity and spatial position.

Sparse Hub Attention preserves a shared interaction pathway while reducing the per-block attention cost from $\mathcal{O}(P^2 n^2 L^2)$ to
\begin{equation}
\mathcal{O}\!\left(P\,nL\,(nL+nK)\right)
+
\mathcal{O}\!\left(nK\,(P\,nL+nK)\right),
\end{equation}
which is linear in $P$ for fixed block size $n$, spatial length $L$, and number of hub tokens $K$.

\subsection{Model Training and Inference}
\label{sec:causal-distillation}
Our deployment target is a conditional few-step causal generator for real-time multi-agent rollout.
A bidirectional diffusion model provides strong visual quality and cross-agent consistency, but cannot be used directly for online generation because it attends to future frames.
A causal model supports KV-cached streaming, but training only on ground-truth histories creates a train--test mismatch during autoregressive rollout.

Inspired by Diffusion Forcing~\cite{chen2024diffusion} and Self-Forcing~\cite{huang2025self}, we therefore use a three-stage recipe tailored to interactive multi-agent simulation: we first train a high-quality bidirectional teacher, then train a block-causal multi-step student with Sparse Hub Attention and Diffusion Forcing, and finally distill the causal student into a conditional few-step generator for low-latency streaming inference.

\noindent\textbf{Bidirectional teacher.}
We first train a bidirectional teacher for high-quality conditional denoising.
The teacher processes the full multi-agent sequence in one forward pass with dense bidirectional attention and a single shared noise level across all agent-time slots.
It is conditioned on the same signals used at inference time, including first-frame observations and per-agent action sequences.
Because the teacher is used only during training, it can exploit full temporal and cross-agent visibility to model local dynamics, agent interactions, and cross-perspective consistency.
This teacher provides the high-quality conditional multi-agent distribution used in the final distillation stage.

\noindent\textbf{Causal student.}
We separately train a causal autoregressive generator using the Diffusion Forcing formulation. 
The student combines block-causal attention from \S\ref{sec:preliminaries} with the Sparse Hub Attention mask $\mathcal{M}$ from \S\ref{sec:sparse-hub-attention}.
Each temporal block receives an independently sampled noise level, and each query attends only to the current or previous blocks.
In addition to this causal constraint, agents communicate through shared hub tokens rather than dense pairwise attention.

Unlike post-training recipes~\cite{huang2025self,yin2025slow} that use causal training mainly as a short warm-up before few-step distillation, we train the causal student as a full multi-step diffusion model. Before any few-step compression, the causal student can already produce reasonable autoregressive rollouts with multi-step sampling, providing a stable starting point for distillation.

\noindent\textbf{Conditional Self-Forcing distillation.}
Finally, inspired by Self-Forcing~\cite{huang2025self}, we distill the multi-step causal student into a few-step generator for real-time inference.
The student is initialized from the trained multi-step causal model, while the bidirectional teacher provides the high-quality conditional distribution-matching signal.
We train the few-step student under autoregressive self-rollout: generated blocks are written to the KV cache and reused as history for subsequent blocks, matching the inference-time rollout distribution.
We apply distribution matching distillation (DMD)~\cite{yin2024one} with rollout-aware training, encouraging the few-step student to preserve quality not only within each block, but also over its own generated histories.

Our setting requires conditional distillation.
Interactive world models must preserve the initial observation and respond to actions, rather than merely produce plausible videos.
We therefore provide the same conditioning package $\mathcal{C}$, including first-frame observations and per-agent actions, to both teacher and student during distillation.
This aligns the conditional rollout distribution and prevents the few-step model from drifting away from the specified initial state or action controls.

\noindent\textbf{KV-cached streaming inference.}
At inference time, the distilled few-step student generates one temporal block at a time, conditioned on the initial observations and the latest per-agent action block, and streams the resulting rollout at 24 FPS.
To preserve the Sparse Hub Attention topology during streaming, we maintain separate KV caches for each agent stream and a shared KV cache for hub tokens.
When generating a new block, each agent reads keys and values from its own past blocks and from the hub cache, while the hub reads cached states from all agents and previous hub tokens.
Thus, cross-agent information still flows only through the hub, even when using cached histories.

\section{Experiments}
\label{sec:experiments}

\begin{table*}[t]
  \centering
  \footnotesize
  \setlength{\tabcolsep}{3.8pt}%
  \renewcommand{\arraystretch}{1.12}%
  \setlength{\belowcaptionskip}{10pt}%
    \caption{Comparison with Solaris across multi-agent evaluation protocols. FID and FVD are lower better ($\downarrow$). Best per column in \textbf{bold}.}%
  \label{tab:method-comparison}%
  \resizebox{0.8\linewidth}{!}{%
    \begin{tabular}{@{}
        >{\raggedright\arraybackslash}p{2.2cm}
        *{10}{c}
        @{}
      }
      \toprule
      \multirow{2}{*}{Method}
        & \multicolumn{2}{c}{Memory}
        & \multicolumn{2}{c}{Grounding}
        & \multicolumn{2}{c}{Movement}
        & \multicolumn{2}{c}{Building}
        & \multicolumn{2}{c}{Consistency} \\
      \cmidrule(lr){2-3} \cmidrule(lr){4-5} \cmidrule(lr){6-7} \cmidrule(lr){8-9} \cmidrule(lr){10-11}
        & FVD $\downarrow$ & FID $\downarrow$
        & FVD $\downarrow$ & FID $\downarrow$
        & FVD $\downarrow$ & FID $\downarrow$
        & FVD $\downarrow$ & FID $\downarrow$
        & FVD $\downarrow$ & FID $\downarrow$ \\
      \midrule
      Frame concat~\cite{enigma2025introducing_multiverse}
        & $450.6$ & $69.8$
        & $528.3$ & $63.2$
        & $556.9$ & $65.0$
        & $551.8$ & $87.3$
        & $576.0$ & $123.2$ \\
      \parbox[t]{2.2cm}{\raggedright Solaris~\cite{savva2026solaris}}
        & $333.8$ & $51.7$
        & $301.9$ & $36.1$
        & $311.1$ & $36.3$
        & $448.6$ & $71.0$
        & $443.1$ & $94.8$ \\
      \textbf{$\gamma$-World (Ours)}
        & \textbf{184.1} & \textbf{24.8}
        & \textbf{199.3} & \textbf{24.0}
        & \textbf{191.5} & \textbf{21.2}
        & \textbf{264.5} & \textbf{32.1}
        & \textbf{280.0} & \textbf{46.9} \\
      \bottomrule
    \end{tabular}%
    
  }%
\end{table*}

\begin{table*}[t]
  \centering
  \footnotesize
  \setlength{\belowcaptionskip}{10pt}%
  \caption{Architecture design. All metrics are averaged over the test scenarios in the game environment.}%
  \label{tab:architecture-ablation}%
  \setlength{\tabcolsep}{3.2pt}%
  \renewcommand{\arraystretch}{1.12}%
  \resizebox{\linewidth}{!}{%
    \begin{tabular}{@{}
        >{\raggedright\arraybackslash}p{2.7cm}
        >{\centering\arraybackslash}p{2.2cm}
        >{\centering\arraybackslash}p{2.0cm}
        >{\centering\arraybackslash}p{1.7cm}
        *{5}{c}
        @{}
      }
      \toprule
      \textbf{Setting}
        & \textbf{Composition}
        & \textbf{Agent Encoding}
        & \textbf{Interaction}
        & FVD $\downarrow$
        & FID $\downarrow$
        & LPIPS $\downarrow$
        & PSNR $\uparrow$
        & SSIM $\uparrow$ \\
      \midrule
      Spatial Concat
        & Spatial concat & None & Full
        & $312.4$ & $38.7$ & $0.326$ & $24.8$ & $0.782$ \\
      Sequence Concat
        & Sequence concat & None & Full
        & $285.6$ & $35.2$ & $0.298$ & $25.6$ & $0.798$ \\
      View Embedding
        & Sequence concat & View emb. & Full
        & $256.3$ & $32.4$ & $0.281$ & $26.4$ & $0.815$ \\
      Simplex Encoding
        & Sequence concat & Simplex & Full
        & $228.5$ & \textbf{29.6} & \textbf{0.265} & $27.5$ & $0.830$ \\
      \textbf{$\gamma$-World (Full)}
        & Sequence concat & Simplex & Sparse Hub
        & \textbf{223.4} & $30.2$ & $0.269$ & \textbf{27.7} & \textbf{0.836} \\
      \bottomrule
    \end{tabular}%
  }%
\end{table*}

\begin{figure}[t]
  \centering
  \includegraphics[width=\linewidth,trim={3pt 8pt 3pt 7pt},clip]{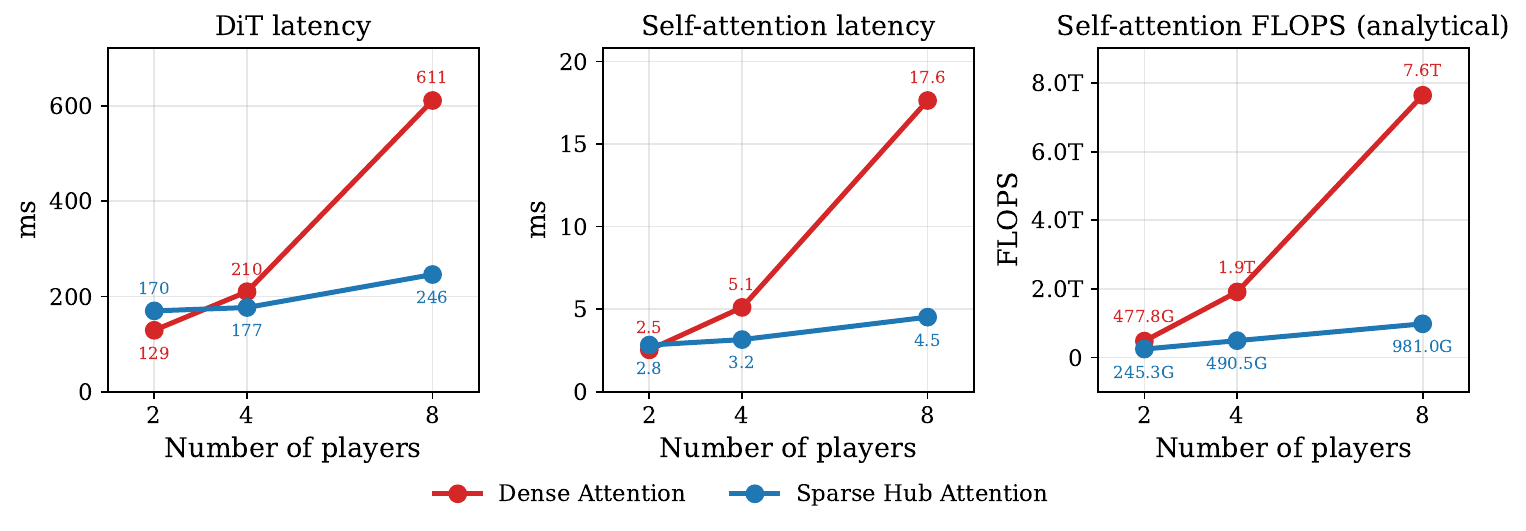}
  \caption{Efficiency comparison between dense cross-agent attention and Sparse Hub Attention across 2, 4, and 8 agents. Sparse Hub Attention achieves significantly lower latency and FLOPs as the number of agents increases.}
  \label{fig:sparse-hub-efficiency}
\end{figure}

\begin{figure*}[t]
  \centering
  \includegraphics[width=\linewidth]{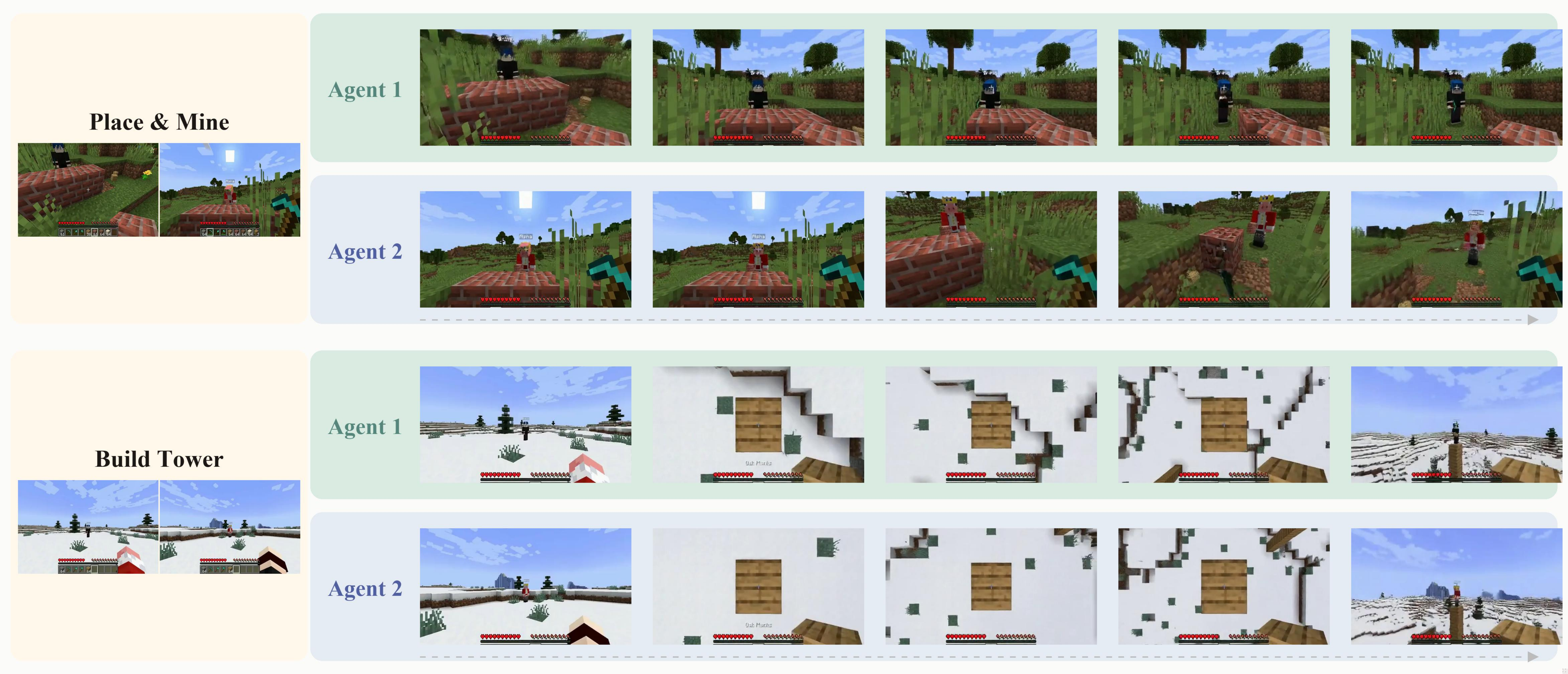}
  \caption{Qualitative examples of two-agent interaction. Each row shows a different task. }
  \label{fig:qualitative-two-agent}
\end{figure*}

\textbf{Experimental setup.} Implementation details are provided in Appendix~\ref{app:additional_implementation_details}.
For virtual-game environments, we construct synchronized multi-agent Minecraft trajectories with a data-generation pipeline inspired by SolarisEngine~\cite{savva2026solaris}, using controllable episode scripts, coordinated bots, and aligned visual-action recording.
The game dataset includes two-agent episodes as the main setting and extends the same collection pipeline to more agents, such as four-agent scenes, enabling evaluation of both pairwise interaction and scalable multi-agent generation.
We evaluate generation quality with FVD and FID, and measure perceptual and pixel-level quality with LPIPS, PSNR, and SSIM.
To evaluate scalability, we measure DiT latency, self-attention latency, and self-attention FLOPs as the number of agents changes.

\subsection{Quantitative Results}
\noindent\textbf{Comparison with multi-agent baselines.}
We compare \ours against two representative baselines for multi-agent world modeling.
The first is a frame-concatenation baseline, following the Multiverse-style design~\cite{enigma2025introducing_multiverse}, which merges multiple agent views into a single visual stream.
The second is Solaris~\cite{savva2026solaris}, a multiplayer Minecraft world model that explicitly trains on synchronized player trajectories.
As shown in Table~\ref{tab:method-comparison}, our method achieves the best overall performance across the evaluation categories, demonstrating the effectiveness of our framework, including order-free agent identity encoding and sparse hub-based interaction.
Compared with frame concatenation, our model avoids compressing multiple agents into a single undifferentiated visual stream, which leads to more coherent interaction modeling and better preservation of each agent's viewpoint.
Compared with Solaris, our design further strengthens agent-level correspondence and information exchange, yielding more reliable results in scenarios that require memory, grounding, building, and cross-view consistency.
These results indicate that multi-agent world modeling benefits from treating agents as distinct but coupled entities rather than simply aggregating their visual observations.

\noindent\textbf{Architecture ablations.}
In Table~\ref{tab:architecture-ablation}, we ablate the main architectural choices in \ours: input organization, agent identity encoding, and cross-agent interaction.
For input organization, \emph{Spatial Concat} merges agent views into a larger canvas, while \emph{Sequence Concat} preserves each agent as a separate stream.
Although these designs can behave similarly in the two-agent setting, spatial concatenation becomes increasingly expensive as the number of agents grows because it increases the effective spatial resolution.
Sequence concatenation keeps the per-agent spatial resolution fixed and is therefore more compatible with variable agent counts.
Given sequence concatenation, we compare learned \emph{View Embedding} against our \emph{Simplex Rotary Agent Encoding}.
Simplex Rotary Agent Encoding improves over learned view embeddings by assigning distinct agent identities without imposing a privileged slot order, matching the exchangeability of agents in a shared world.
Finally, we replace dense cross-agent attention with \emph{Sparse Hub Attention} to test whether agents can communicate through a compact shared state.
As shown in Table~\ref{tab:architecture-ablation}, the full design, combining sequence-based multi-agent tokenization, Simplex Rotary Agent Encoding, and Sparse Hub Attention, achieves strong visual quality and consistency across FVD, FID, LPIPS, PSNR, and SSIM.
These ablations support our core design principle: agents should be represented as distinct but exchangeable entities and coupled through an efficient shared interaction pathway.

\noindent\textbf{Efficiency of Sparse Hub Attention.}
We further evaluate the efficiency of \emph{Sparse Hub Attention} by comparing it with full cross-agent attention as the number of agents increases.
This comparison directly tests the motivation in Sec.~\ref{sec:sparse-hub-attention}: dense all-to-all interaction grows quadratically with the number of agents, whereas \emph{Sparse Hub Attention} routes cross-agent information through a compact shared state.
We report DiT latency, self-attention latency, and self-attention FLOPs in Figure~\ref{fig:sparse-hub-efficiency}. Latency is averaged over 3 full rollouts to 24 latent frames with full KV cache, and FLOPs is computed analytically from the average token sequence length over the 24-latent rollout.
\emph{Sparse Hub Attention} substantially reduces computation time and FLOPs as the number of agents grows, while dense attention quickly becomes expensive at larger agent counts.
These results show that \emph{Sparse Hub Attention} provides a practical scaling path for multi-agent world models beyond two players.

\subsection{Qualitative Results}
We complement the quantitative evaluation with qualitative examples that visualize how the model preserves shared-world consistency across multiple agent viewpoints.

\noindent\textbf{Two-agent interaction.}
Figure~\ref{fig:qualitative-two-agent} shows two-agent rollouts generated by \ours.
The paired streams remain synchronized: actions taken by one agent are reflected in the other agent's observation when the agents interact in the shared environment.
The model also maintains object and agent grounding when agents temporarily move out of each other's field of view, suggesting that it tracks a shared latent world state rather than generating independent single-agent videos.

\begin{figure*}[t]
  \centering
  \includegraphics[width=\linewidth]{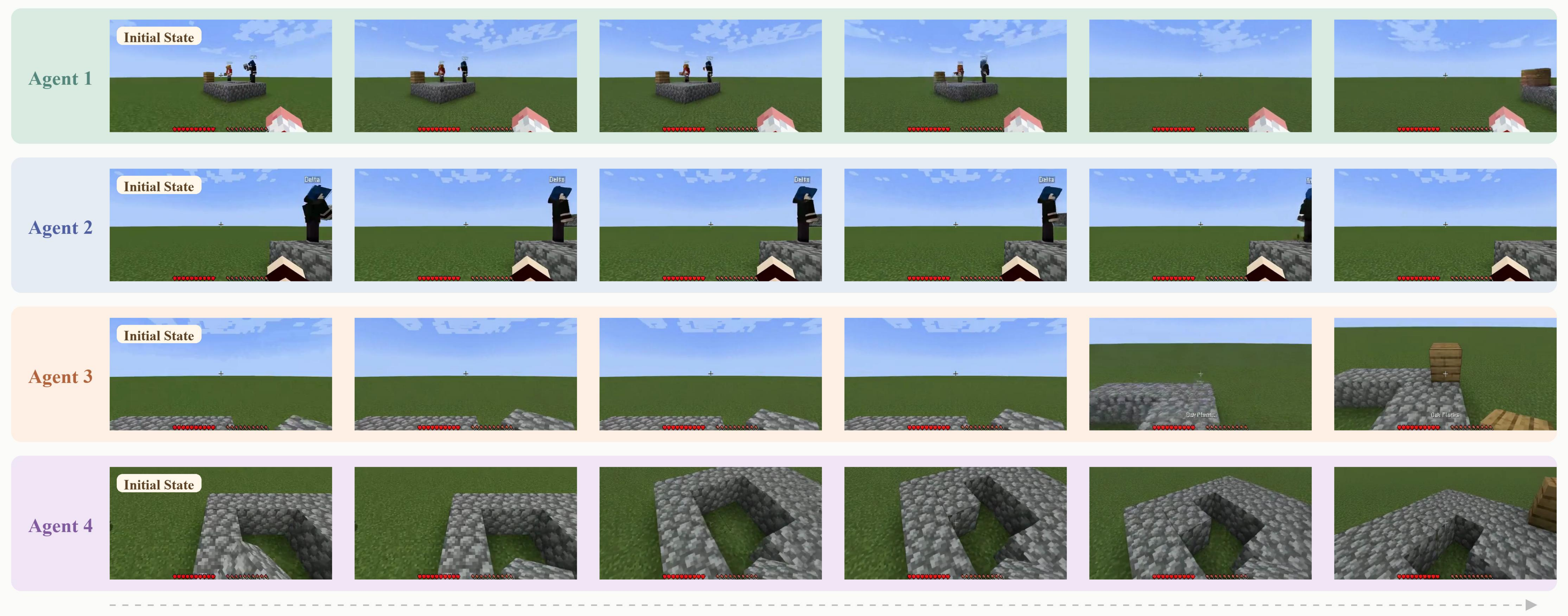}
  \caption{Qualitative example of scaling beyond two players. The first frame in each row shows the initial state for one agent, followed by synchronized rollouts across four agents.}
  \label{fig:qualitative-scaling}
\end{figure*}

\noindent\textbf{Scaling beyond two players.}
Figure~\ref{fig:qualitative-scaling} shows zero-shot four-agent rollouts from a model trained only on two-agent data.
The same model generates synchronized visual streams for multiple players without changing the architecture.
This scaling behavior is enabled by Simplex Rotary Agent Encoding, which avoids fixed learned slot identities, and Sparse Hub Attention, which provides a shared communication pathway without dense pairwise attention.
The examples indicate that \ours learns coupled multi-agent dynamics rather than independently rolling out each view.

\begin{figure*}[t]
  \centering
  \includegraphics[width=\linewidth]{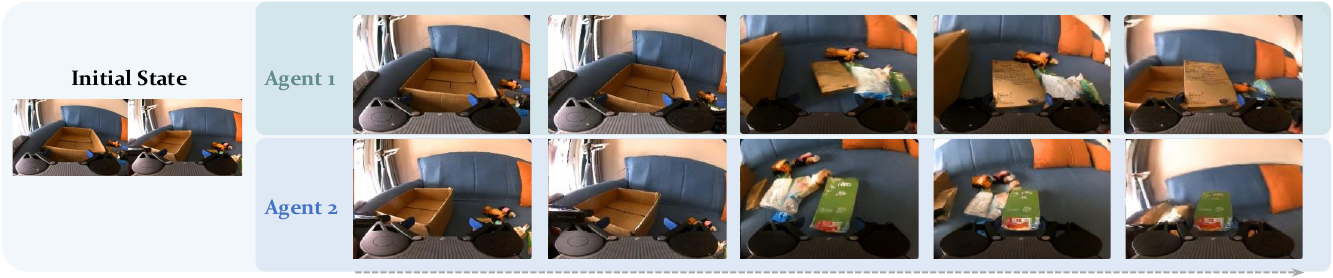}
  \caption{Qualitative examples of real-world robotic coordination.}
  \label{fig:real-world-robotics}
\end{figure*}

\noindent\textbf{Real-world robotics applications.}
Beyond virtual game environments, we also evaluate \ours on real-world robotics coordination tasks.
We use the RealOmin-Open Dataset~\cite{genrobot2025opendata}, treating the left and right robot arms as two interacting agents---this lets the same multi-agent world-modeling framework that handles virtual players in Minecraft also capture coordinated bimanual manipulation in physical scenes.
As shown in Figure~\ref{fig:real-world-robotics}, the model generates future frames that preserve the coordinated motion of multiple robotic agents and the spatial layout of the scene.
These examples indicate that the same generative multi-agent formulation can extend from virtual games to real-world environments.

\section{Discussion}
We presented \ours, a generative multi-agent world model for interactive simulation beyond two players.
\ours combines Simplex Rotary Agent Encoding for distinct yet permutation-symmetric agent identities with Sparse Hub Attention for efficient hub-mediated cross-agent communication.
Together with conditional teacher--student distillation and KV-cached streaming inference, these components enable real-time action-responsive rollouts that remain consistent across time and agent perspectives.
Experiments in multiplayer virtual environments show that \ours improves fidelity, controllability, and inter-agent consistency over baselines.
Ablations validate the design choices, while scaling experiments show that the model generalizes from two-player training to four-player simulation without additional training and achieves lower inference cost as the number of agents increases.
We further demonstrate the same formulation on real-world robotic coordination by treating the left and right robot arms as interacting agents.

\noindent\textbf{Limitations.}
Our current evaluation focuses primarily on gaming environments and robotics examples; broader validation in more complex, heterogeneous, and long-horizon settings remains future work.
The simplex pool supports agent-count scaling within a fixed rotary agent band, but very large populations may require larger bands or hierarchical agent grouping. Finally, because \ours does not explicitly enforce 3D geometry or physical constraints, long rollouts may still accumulate inconsistencies.

\section*{Acknowledgements}
The authors would like to thank Product Managers Aditya Mahajan and Matt Cragun for their valuable support and guidance, Jingnan Gao for proof discussion, and Yixin Hong for demo creation.

\clearpage
\newpage
\appendix
\section{More Visualizations}
We include extended 24-second multi-agent rollouts, qualitative comparisons against baselines, and real-world robotic coordination videos in the supplementary video, offering a more intuitive demonstration of our method.

\section{Proof of Simplex Equidistance}
\label{app:simplex-equidistance}

We provide the derivation for the equidistance property of Simplex Rotary Agent Encoding. 
Let $V$ denote the simplex pool size, matching the notation of the main paper.
We first construct the simplex vertices in $\mathbb{R}^{V}$ and then embed them into the $(d_p/2)$-dimensional agent angle space, where $d_p$ is the agent rotary band size and $V\leq d_p/2+1$.

Let $\mathbf{e}_p\in\mathbb{R}^{V}$ denote the one-hot vector of agent $p$, and let $\mathbf{1}\in\mathbb{R}^{V}$ denote the all-one vector. 
We define the centered vector
\begin{equation}
\bar{\mathbf{s}}_p
=
\mathbf{e}_p-\frac{1}{V}\mathbf{1}.
\end{equation}
These vectors lie in the zero-mean subspace because
\begin{equation}
\mathbf{1}^{\top}\bar{\mathbf{s}}_p=0.
\end{equation}
For any agent $p$, the squared norm of $\bar{\mathbf{s}}_p$ is
\begin{align}
\|\bar{\mathbf{s}}_p\|_2^2
&=
\left(
\mathbf{e}_p-\frac{1}{V}\mathbf{1}
\right)^{\top}
\left(
\mathbf{e}_p-\frac{1}{V}\mathbf{1}
\right) \\
&=
1-\frac{2}{V}+\frac{1}{V} \\
&=
\frac{V-1}{V}.
\end{align}
For two distinct agents $p\neq q$, their inner product is
\begin{align}
\bar{\mathbf{s}}_p^{\top}\bar{\mathbf{s}}_q
&=
\left(
\mathbf{e}_p-\frac{1}{V}\mathbf{1}
\right)^{\top}
\left(
\mathbf{e}_q-\frac{1}{V}\mathbf{1}
\right) \\
&=
0-\frac{1}{V}-\frac{1}{V}+\frac{1}{V} \\
&=
-\frac{1}{V}.
\end{align}
After normalization, we obtain
\begin{equation}
\mathbf{s}_p
=
\sqrt{\frac{V}{V-1}}
\bar{\mathbf{s}}_p.
\end{equation}
Therefore,
\begin{align}
\|\mathbf{s}_p\|_2^2 &= 1, \\
\mathbf{s}_p^{\top}\mathbf{s}_q
&=
-\frac{1}{V-1},
\qquad p\neq q.
\end{align}
The squared distance between any two distinct simplex vertices ($V\geq 2 $) is then
\begin{align}
\|\mathbf{s}_p-\mathbf{s}_q\|_2^2
&=
\|\mathbf{s}_p\|_2^2
+
\|\mathbf{s}_q\|_2^2
-
2\mathbf{s}_p^{\top}\mathbf{s}_q \\
&=
2+\frac{2}{V-1} \\
&=
\frac{2V}{V-1},
\qquad p\neq q.
\end{align}
Thus, all distinct agent pairs are exactly equidistant in the simplex angle space.

In Simplex Rotary Agent Encoding, the agent angle is defined as
\begin{equation}
\boldsymbol{\theta}_p
=
\alpha\mathbf{s}_p,
\end{equation}
where $\alpha$ is a scalar scale factor. 
Therefore, for any $p\neq q$,
\begin{align}
\|\boldsymbol{\theta}_p-\boldsymbol{\theta}_q\|_2^2
&=
\alpha^2
\|\mathbf{s}_p-\mathbf{s}_q\|_2^2 \\
&=
\alpha^2
\frac{2V}{V-1}.
\end{align}
Hence, all different-agent pairs receive the same amount of separation in the underlying agent angle space. We further analyze the induced distance in the complex RoPE space. 
The complex rotary representation of agent $p$ is
\begin{equation}
\boldsymbol{\Phi}_p
=
\exp(i\boldsymbol{\theta}_p).
\end{equation}
For two agents $p$ and $q$, the squared distance between their complex rotations is
\begin{align}
\|\boldsymbol{\Phi}_p-\boldsymbol{\Phi}_q\|_2^2
&=
\left\|
\exp(i\boldsymbol{\theta}_p)
-
\exp(i\boldsymbol{\theta}_q)
\right\|_2^2 \\
&=
\sum_{r=1}^{d_p/2}
\left|
e^{i\theta_p^r}
-
e^{i\theta_q^r}
\right|^2 \\
&=
\sum_{r=1}^{d_p/2}
2\left(
1-\cos(\theta_p^r-\theta_q^r)
\right).
\end{align}
For sufficiently small $\alpha$ such that the coordinate-wise angle differences $|\theta_p^r-\theta_q^r|$ are small for all $r$, we use the approximation $1-\cos x\approx x^2/2$. 
Thus,
\begin{align}
\|\boldsymbol{\Phi}_p-\boldsymbol{\Phi}_q\|_2^2
&\approx
\sum_{r=1}^{d_p/2}
(\theta_p^r-\theta_q^r)^2 \\
&=
\|\boldsymbol{\theta}_p-\boldsymbol{\theta}_q\|_2^2 \\
&=
\alpha^2
\frac{2V}{V-1},
\qquad p\neq q.
\end{align}
This shows that Simplex Rotary Agent Encoding guarantees exact equidistance in the agent angle space and yields approximately equal pairwise separation in the complex RoPE space. 
In the common implementation where the centered one-hot simplex is embedded with zero padding and $d_p/2\geq V$, the complex-space distance is also identical across distinct agent pairs because all pairwise difference vectors share the same non-zero coordinate pattern up to permutation.

\section{More Architecture}
\label{app:more-architecture}

\subsection{Action Design}
\label{app:action-design}

We use explicit per-agent action traces as additional conditioning signals.
For both domains, actions are synchronized with the video frames and provided separately for different agents.
The action specification is domain-specific: the game domain uses player control commands, while the robot domain uses continuous end-effector state.

\noindent\textbf{Game action format.}
For Minecraft-style game videos, each agent action at a frame contains $25$ fields: $23$ discrete player controls and $2$ continuous camera controls.
The discrete controls cover inventory interaction, hotbar selection, movement, item manipulation, and mouse-button actions.
The continuous camera controls describe horizontal and vertical view motion.
Table~\ref{tab:game-action-format} summarizes the game action fields.

\begin{table*}[t]
  \centering
  \footnotesize
  \caption{Game action format. Each frame stores a $25$-field action vector for each agent.}
  \label{tab:game-action-format}
  \vspace{4pt}
  \setlength{\tabcolsep}{4pt}
  \renewcommand{\arraystretch}{1.12}
  \begin{tabular}{@{}lll@{}}
    \toprule
    \textbf{Index} & \textbf{Field} & \textbf{Description} \\
    \midrule
    $0$ & \texttt{inventory} & Open inventory \\
    $1$ & \texttt{ESC} & Exit or cancel current menu \\
    $2$--$10$ & \texttt{hotbar.1}--\texttt{hotbar.9} & Select hotbar slot 1--9 \\
    $11$--$14$ & \texttt{forward}, \texttt{back}, \texttt{left}, \texttt{right} & Locomotion controls \\
    $15$--$17$ & \texttt{jump}, \texttt{sneak}, \texttt{sprint} & Movement modifiers \\
    $18$ & \texttt{swapHands} & Swap held item between hands \\
    $19$--$22$ & \texttt{attack}, \texttt{use}, \texttt{pickItem}, \texttt{drop} & Interaction and item manipulation \\
    $23$--$24$ & \texttt{cameraX}, \texttt{cameraY} & Horizontal yaw and vertical pitch motion \\
    \bottomrule
  \end{tabular}
\end{table*}

\noindent\textbf{Robot action format.}
For robot videos, each agent action at a frame contains $10$ continuous fields describing the robot end-effector and gripper.
The action contains the 3D end-effector position, a 6D orientation field, and the gripper opening value.
The same format is used for the left and right robots, producing one temporally aligned action sequence per robot.
Table~\ref{tab:robot-action-format} summarizes the robot action fields.

\begin{table}[t]
  \centering
  \footnotesize
  \caption{Robot action format. Each frame stores a $10$-field continuous action vector for each robot.}
  \label{tab:robot-action-format}
  \vspace{4pt}
  \setlength{\tabcolsep}{4pt}
  \renewcommand{\arraystretch}{1.12}
  \begin{tabular}{@{}lll@{}}
    \toprule
    \textbf{Index} & \textbf{Field} & \textbf{Description} \\
    \midrule
    $0$--$2$ & \texttt{pos\_x}, \texttt{pos\_y}, \texttt{pos\_z} & End-effector position \\
    $3$--$8$ & \texttt{rot\_6d\_0}--\texttt{rot\_6d\_5} & End-effector orientation \\
    $9$ & \texttt{gripper} & Gripper opening value \\
    \bottomrule
  \end{tabular}
\end{table}

\section{Additional Implementation Details}
\label{app:additional_implementation_details}

\noindent\textbf{Architecture.}
Both bidirectional teacher model and causal student model are based on Cosmos-Predict2.5-2B~\cite{ali2025world-simulation},  with hidden dimension $D=2048$, $28$ transformer blocks, $16$ attention heads (head dimension $128$), MLP ratio $4$, and AdaLN-LoRA of rank $256$. Player identity is injected via our Simplex Rotary Agent Encoding (Sec.~\ref{sec:simplex-rotary-agent-encoding}), partitioning the head dimension as $(64, 32, 16, 16)$ across the $(t, p, h, w)$ axes and assigning each agent a vertex of a regular simplex as its rotation phase. We use a simplex pool of size $4$ over $2$ active runtime slots: at every training step we randomly sample $2$ of the $4$ vertices and additionally permute the agent slot order, forcing the model to disambiguate players exclusively through the simplex marker and allowing the same checkpoint to serve up to $4$ players at inference. Per-player actions (a $23$-dimensional keyboard one-hot and a $2$-dimensional camera vector) are encoded by a single shared action encoder: two MLP branches that lift each modality to $128$ dimensions, a fusion MLP that maps the concatenated $256$-dimensional vector through a $4\times$-stride 1D temporal convolution, and a final projection to the model dimension $D=2048$. The encoder output is added as a per-block bias to the self-attention input of every transformer layer. The causal student additionally enables (i) \emph{Sparse Hub Attention}: each player attends only to its own past tokens and to $K=8$ learnable global hub tokens per latent frame; and (ii) \emph{local windowed attention}: each query attends to the most recent $24$ latent frames per view, bounding the inference KV cache independently of generation length. The bidirectional teacher uses standard dense self-attention.

\noindent\textbf{Stages 1 and 2 -- Bidirectional teacher and causal student pretraining.}
Both the teacher and the student are initialized from the publicly released Cosmos-Predict2.5-2B TI2V checkpoint and trained on 2-agent gameplay at $320 \times 480$ per view. The teacher is first pretrained on $93$-frame clips (latent length $24$) for $10{,}000$ iterations and then fine-tuned on $189$-frame clips (latent length $48$) for $6{,}000$ iterations, while the causal student is pretrained on $93$-frame clips for $15{,}000$ iterations. Both stages use AdamW with learning rate $3{\times}10^{-5}$, weight decay $10^{-3}$, $(\beta_1, \beta_2) = (0.0, 0.999)$, $100$-step linear warm-up, and gradient clipping at $0.1$. The teacher is trained on $32$ NVIDIA GB200s and the student on $32$ NVIDIA GB200s. We do not apply classifier-free guidance (CFG) at inference, as we empirically observe that unguided sampling produces more accurate results.

\noindent\textbf{Stage 3 -- Self-Forcing distillation.}
The distillation model couples three networks: the student (trainable, initialized from Stage 2), a frozen \emph{real score} (the Stage-1 teacher), and a trainable \emph{fake score} also initialized from the Stage-1 teacher. We follow Self-Forcing~\cite{huang2025self} to optimize the student model with the DMD~\cite{yin2024one} loss on $189$-frame clips. At each generator step, each block is denoised with timesteps $\{1000, 750, 500, 250\}$ (warped by the flow shift $5.0$). After each block, the model re-forwards the block under context-noise level $128$ and writes the result into the per-layer KV cache before proceeding. Generator and critic are updated alternately at a $1{:}4$ ratio: the student is updated once every $5$ iterations, while on the remaining iterations the fake critic is updated with a flow-velocity prediction loss; the real score is kept frozen throughout. Distillation runs for $400$ iterations on $32$ NVIDIA GB200s, using AdamW with learning rates $2{\times}10^{-6}$ (student) and $4{\times}10^{-7}$ (critic), weight decay $10^{-2}$, $(\beta_1, \beta_2) = (0.0, 0.999)$.

\noindent\textbf{Inference.}
At inference time, the student generates each per-view latent block autoregressively with the same $4$-step denoising schedule and block size as in training. The KV cache uses the rolling local-attention window of $24$ latent frames per view, decoupling the generated sequence length from cache memory.

\section{Additional Ablations}
\label{app:additional-ablations}

We compare three variants of our model: causal, bidirectional, and distilled.
As shown in Table~\ref{tab:model-variant-ablation}, we report FVD, FID, LPIPS, PSNR, and SSIM to evaluate perceptual quality and reconstruction fidelity.

\begin{table}[t]
  \centering
  \footnotesize
  \caption{Comparison of causal, bidirectional, and distilled variants of our model.}
  \label{tab:model-variant-ablation}
  \vspace{4pt}
  \setlength{\tabcolsep}{4pt}
  \renewcommand{\arraystretch}{1.12}
  \begin{tabular}{@{}lccccc@{}}
    \toprule
    \textbf{Variant}
      & FVD $\downarrow$
      & FID $\downarrow$
      & LPIPS $\downarrow$
      & PSNR $\uparrow$
      & SSIM $\uparrow$ \\
    \midrule
    Bidirectional
      & $227.3$ & $31.0$ & $0.272$ & $27.7$ & $0.828$ \\
    Causal
      & $266.4$ & $34.4$ & $0.277$ & $26.2$ & $0.805$ \\
    Distilled
      & $239.7$ & $30.9$ & $0.273$ & $26.8$ & $0.811$ \\
    \bottomrule
  \end{tabular}
\end{table}

We provide an additional ablation on the number of hub tokens $K$ used in Sparse Hub Attention.
In our design, hub tokens serve as the compact shared state through which agents exchange information, so $K$ controls the capacity of the cross-agent communication bottleneck.
When $K$ is too small, the hub has limited capacity to summarize multi-agent state, which can hurt generation quality.
Increasing $K$ provides a richer shared state by allowing the hub to summarize multi-agent interactions with greater capacity.
As shown in Table~\ref{tab:hub-token-ablation}, we sweep $K\in\{1,8,32,128\}$ and report FVD, FID, LPIPS, PSNR, and SSIM.
This ablation helps identify a practical operating point where the hub has enough capacity for cross-agent interaction.

\begin{table}[t]
  \centering
  \footnotesize
  \setlength{\belowcaptionskip}{10pt}%
  \caption{Ablation on the number of hub tokens in Sparse Hub Attention. We vary the hub token count $K$ and report generation quality, perceptual quality, and pixel-level quality.}%
  \label{tab:hub-token-ablation}%
  \vspace{4pt}%
  \setlength{\tabcolsep}{4pt}%
  \renewcommand{\arraystretch}{1.12}%
    \begin{tabular}{@{}lccccc@{}}
    \toprule
    \textbf{Hub Tokens ($K$)}
      & FVD $\downarrow$
      & FID $\downarrow$
      & LPIPS $\downarrow$
      & PSNR $\uparrow$
      & SSIM $\uparrow$ \\
    \midrule
    $1$
      & $250.9$ & $31.5$ & $0.271$ & $27.3$ & $0.825$ \\
    $8$
      & $223.4$ & $30.2$ & $0.269$ & $27.7$ & $0.836$ \\
    $32$
      & $221.8$ & $29.8$ & $0.267$ & $27.9$ & $0.838$ \\
    $128$
      & $220.5$ & $29.5$ & $0.266$ & $28.0$ & $0.839$ \\
    \bottomrule
  \end{tabular}
\end{table}

\section{More Experimental Results}
\label{app:more-experimental-results}

\subsection{Training Stage Comparison}
\label{app:training-stage-comparison}

We compare three training stages of our model: the bidirectional teacher, the causal student, and the distilled model.
As shown in Table~\ref{tab:model-variant-ablation}, the bidirectional teacher achieves the best overall performance due to its access to full temporal context during generation.
The causal variant, while enabling streaming inference with KV caching, shows degraded performance because it can only attend to past frames.
The distilled model recovers much of the teacher's quality while retaining the causal structure, demonstrating that knowledge distillation effectively transfers bidirectional modeling capacity into a streaming-compatible architecture.

\clearpage
{
\small
\bibliographystyle{abbrv}
\bibliography{main}
}
\end{document}